\documentclass[11pt]{article}

\usepackage[preprint]{acl}

\usepackage{times}
\usepackage{latexsym}
\usepackage[T1]{fontenc}
\usepackage[utf8]{inputenc}
\usepackage{microtype}
\usepackage{inconsolata}
\usepackage{graphicx}
\usepackage{booktabs}
\usepackage{multirow}
\usepackage{amsmath}
\usepackage{url}

\newcommand{\systemname}{\mbox{EviDAG}}

\title{\systemname{}: Auditable Causal DAG Authoring with Biomedical Literature}

\author{
  \textbf{Yi-han Sheu\textsuperscript{1,2,*}},
  \textbf{Michael R. Steigman\textsuperscript{1}},
  \textbf{Yu Zhou\textsuperscript{1}},
  \textbf{Bo Wang\textsuperscript{1,2}},
  \textbf{Fan-Yu Yen\textsuperscript{3}},
  \textbf{Jordan W. Smoller\textsuperscript{1,2}} \\
  \textsuperscript{1}Massachusetts General Hospital, USA \\
  \textsuperscript{2}Harvard University, USA \\
  \textsuperscript{3}Northeastern University, USA \\
  \small{\textsuperscript{*}\texttt{ysheu@mgh.harvard.edu}}
}

\begin{document}
\maketitle
\begin{abstract}
Constructing causal directed acyclic graphs (DAGs) is a core step in biomedical causal analysis, yet it remains a largely manual process. Analysts must connect study variables to prior literature, evaluate uncertain causal claims, and preserve sufficient provenance for expert review. We present \systemname{}, a browser-based system for authoring causal DAGs as auditable, evidence-linked artifacts from biomedical literature. Given free-text descriptions of study concepts, \systemname{} creates a reproducible literature snapshot, uses an LLM-based reasoning module to generate structured pairwise causal judgments, links literature-supported judgments to verbatim evidence excerpts, and assembles the judgments into a constraint-checked graph. Each proposed edge includes confidence estimates, provenance, and a reviewable rationale. The interface supports study specification, progress monitoring, evidence review, graph comparison, adjustment-set computation, and export. In evaluations against both compact benchmark DAGs and reference DAGs derived from published literature, \systemname{} achieves high edge recall on the literature-based cohort while retaining verifiable evidence trails absent from LLM-only baselines. \systemname{} thus reduces the burden of causal DAG curation while making the resulting assumptions auditable, supporting the design, analysis, and interpretation of biomedical studies.\footnote{The asterisk in the author list denotes the corresponding author. PubMed metadata, Unified Medical Language System (UMLS)-derived fields, and reference DAGs retain their source licenses and access terms.}

\end{abstract}

\section{Introduction}
\label{sec:introduction}

Causal directed acyclic graphs (DAGs) make causal assumptions explicit in biomedical
research, helping researchers reason about confounding, mediation, selection, and adjustment before
fitting statistical models~\citep{greenland1999causal,pearl2009causality}. Constructing a DAG for a
concrete biomedical question, however, remains largely manual. Analysts must combine domain knowledge
with evidence from a large and noisy literature,
decide whether each variable pair is plausibly causal, associational, or unrelated, and preserve enough
provenance for collaborators or reviewers to inspect the graph. This burden grows quadratically with the
number of variables, and the resulting artifacts are often difficult to reproduce because the graph is
separated from the search queries, article snapshots, quoted evidence, and rationale that motivated each
edge.

Large language models (LLMs) can synthesize heterogeneous evidence and reason over natural-language
descriptions of biomedical variables, but directly prompting an LLM to produce a DAG leaves users unable
to audit which claims came from literature, ontology relations, modeling priors, or user constraints.
Current LLM-assisted causal-graph work often uses LLMs as causal-reasoning modules, full-graph discovery
agents, or sources of graph priors \citep{kiciman2023causal,darvariu2024llmpriors,jiralerspong2024efficient}.
For scientific use, such automation needs traceability, review, and explicit failure modes.

We present \systemname{}, a browser-based system for auditable causal DAG authoring from biomedical study
concepts, with pairwise causal judgments and graph assembly informed by retrieved literature evidence,
model prior knowledge, and optional ontology enrichment and researcher-supplied context. \systemname{}
treats DAG construction as a staged
workflow: it normalizes free-text variables with optional UMLS enrichment, threads study context into
concept parsing, causal reasoning, and graph assembly, constructs reproducible PubMed corpus snapshots,
synthesizes evidence-grounded pairwise causal judgments, and uses an LLM-assisted constrained graph
assembly module to produce a DAG under acyclicity, temporality, exogeneity, and user-specified
constraints. The output includes a primary DAG, evidence cards, ambiguity-driven alternatives when
applicable, qualitative structural forms, on-demand adjustment-set artifacts, and exports for downstream
causal analysis.

\systemname{} is designed for biomedical researchers, epidemiologists, clinical investigators, biomedical
informaticians, and causal inference practitioners drafting and reviewing literature-based causal
assumptions. It turns literature review, pairwise causal assessment, and graph assembly into an
inspectable workflow while letting experts accept, reject, or revise machine suggestions. It may also
interest NLP researchers studying
provenance-instrumented, benchmarkable LLM workflows whose outputs remain traceable to sources and
reference graphs.

The system contribution is threefold. First, \systemname{} integrates ontology-aware concept resolution,
literature retrieval, evidence-grounded LLM judgment, and constrained graph solving in a single pipeline,
treating explicit temporal ordering as a first-class constraint in pairwise reasoning and DAG assembly.
Second, it exposes this pipeline through a web interface for submission, progress inspection, evidence
review, graph comparison, and export. Third, it includes a headless mode for reproducible evaluation
against reference DAGs, complementing the interactive workflow.

The remainder describes the architecture, interface, and evaluation. We evaluate \systemname{} on compact
synthetic and published literature DAGs using pairwise and graph-level metrics, and quantify evidence
grounding by citation verifiability and the share of judgments backed by retrieved literature rather than
model prior.

\section{System Architecture}
\label{sec:system-architecture}

\begin{figure*}[t]
\centering
\includegraphics[width=\textwidth]{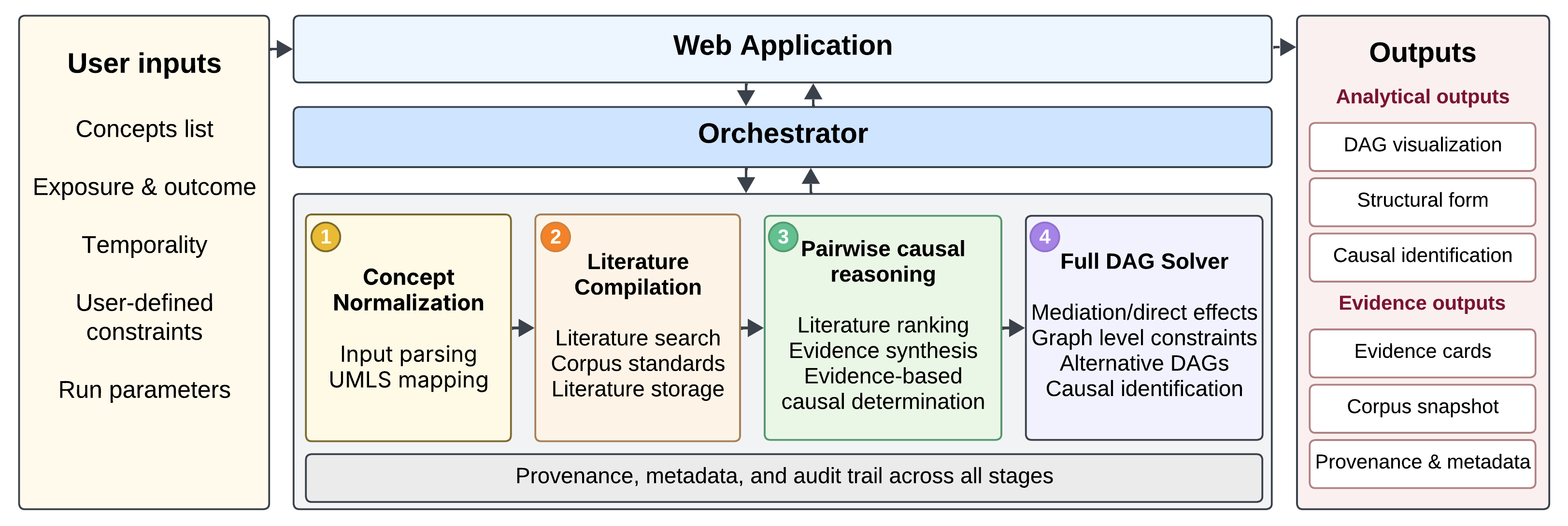}
\caption{\systemname{} system flow. LLM calls occur inside bounded backend stages; the interface renders persisted artifacts rather than recomputing causal state.}
\label{fig:system-flow}
\end{figure*}

\systemname{} is a web interface over a staged backend pipeline. The interface submits requests, renders
persisted artifacts, and exposes controls for submission, review, export, and evaluation, but performs
no causal reasoning itself. LLM calls, biomedical-resource queries, and graph-constraint operations sit
behind typed backend contracts. A central orchestrator owns stage sequencing, persistence, review
pauses, headless execution, and typed artifact handoff across concept resolution, literature search,
pairwise causal reasoning, and graph assembly. The same orchestrator drives batches launched from the
evaluation console, running the pipeline for each selected benchmark case and forwarding completed
outputs to the evaluation backend for scoring against reference DAGs. Figure~\ref{fig:system-flow}
summarizes this boundary.

\paragraph{Terminology.}
A \emph{concept} is a submitted study variable resolved into the normalized random-variable identity
that becomes a graph node; a \emph{concept pair} is an unordered pair of resolved concepts. A
\emph{study context} is optional free text about the population, design, measurement, or variable
definitions that should inform curation without overriding retrieved evidence or hard constraints. A
\emph{stage artifact} is the structured output of one stage, and a \emph{provenance channel} records
whether a claim comes from retrieved literature, UMLS, temporal information, user constraints, solver
decisions, or model prior.

\paragraph{Concept resolution.}
\systemname{} parses each submitted variable string into a core concept representing a graph node's
random-variable identity. It preserves the original string and records qualifiers and directional or
temporal modifiers separately, retaining modifiers when their removal would change the variable's referent
or measurement. Each parsed concept receives a stable local identifier used as the node id even without an
ontology match. Study context informs parsing and candidate selection to disambiguate variable meanings.
When UMLS enrichment is enabled, \systemname{} links the core concept to a Concept Unique Identifier (CUI),
records its preferred name (canonical label), and collects UMLS atoms (source-vocabulary terms) for query
expansion; ambiguous parses and mappings remain reviewable.

\paragraph{Literature search.}
\systemname{} builds a frozen PubMed evidence base for each run by enumerating all unordered concept pairs
and issuing tiered block queries. Query blocks combine submitted terms, parsed names, lemmatized
variants, and available UMLS atoms. Tier 1 requires both concept blocks plus a relationship block
containing causal, associational, effect-size, and null-effect terms; Tier 2 drops the relationship block,
and Tier 3 additionally removes any species restriction. The search stops at the first tier meeting the
retrieval target and logs every attempt. A per-pair cap bounds retrieval, with truncated queries recorded
under PubMed's Best Match ordering. The resulting snapshot includes a deduplicated article catalog,
execution logs, filters, verification summaries, and a search-provenance summary inspired by the Preferred
Reporting Items for Systematic Reviews and Meta-Analyses (PRISMA) \citep{page2021prisma}; downstream
modules consume it read-only.

\paragraph{Pairwise causal reasoning.}
For each concept pair, \systemname{} first ranks evidence by study-design tier, using recency to break ties. It
extracts same-sentence co-occurrences as citable snippets and consults full abstracts for cross-sentence
mentions. Tagged snippets enable precise citations, and the LLM also considers study design, sample size,
methodological quality, and recency. With UMLS relations, optional study context, and model prior knowledge,
this evidence produces a structured judgment specifying relationship type, direction, sign, confidence,
citations, rationale, and per-field provenance. Downstream edges remain traceable.

\paragraph{Graph assembly and identification.}
\systemname{} assembles a graph by combining pairwise causal judgments with optional study context and
user-supplied constraints, such as temporality anchors, exogeneity flags, and required or forbidden edges.
It first translates directional causal judgments into graph edges, while treating associations more
cautiously. Depending on the available evidence and constraints, an association may become a possible
direct edge, suggest a latent common cause, introduce an admitted latent node, or leave the graph unchanged.
Null judgments are recorded for provenance but do not add edges. The assembly process prioritizes
deterministic decisions: \systemname{} applies hard constraints and enforces acyclicity before using a bounded
graph-level LLM choice in cases where local judgments require broader graph context. It also checks whether
proposed direct effects may instead be explained by mediation. Under a lenient policy, \systemname{} retains a
direct edge unless the evidence indicates that indirect paths fully account for the effect; under a
conservative policy, it retains the edge only when a substantial direct effect is supported. \systemname{} then
surfaces unexplained associations as candidate unmeasured common causes.\footnote{An association without a
direct causal effect can reflect an unmeasured common cause or
collider conditioning. \systemname{} surfaces these as candidate unmeasured common causes. Identification
assumes candidates are unconfounded unless users admit selected pairs as latent common causes for on-demand
analysis.}

\systemname{} outputs a primary directed DAG, residual unexplained-association links, alternative DAGs that
highlight edge-level differences, qualitative structural forms, mediation logs, an on-demand
identification workspace, adjustment-set artifacts, evidence logs, and an audit trail. Alternative DAGs
are generated from acyclic one-edge reversals of retained bidirectional-origin edges, making structural
differences visible without recomputing identification. Candidate latent common causes are stored with
rationales; only accepted candidates become explicit unmeasured nodes in the DAG, which can be shown or
hidden in the graph view. \systemname{} computes minimal or maximal backdoor adjustment sets on demand using
DoWhy \citep{sharma2020dowhy}.

\paragraph{Orchestration and audit.}
The orchestrator validates submissions, dispatches stages, persists artifacts, detects review pauses,
and records the submitted configuration, concept rebinding map, module versions, and model versions. The
same orchestration runs headlessly through management commands for continuous integration, benchmarks,
and paper evaluation.

\section{User Interface and Workflow}
\label{sec:interface}

\systemname{}'s browser interface exposes three main workflows: define a study, review its evidence and graph, and
evaluate performance against reference DAGs. It renders persisted artifacts, so every claim traces to a
submission, literature snapshot, pairwise judgment, graph operation, or evaluation bundle.

\begin{figure*}[t]
\centering
\includegraphics[width=\textwidth]{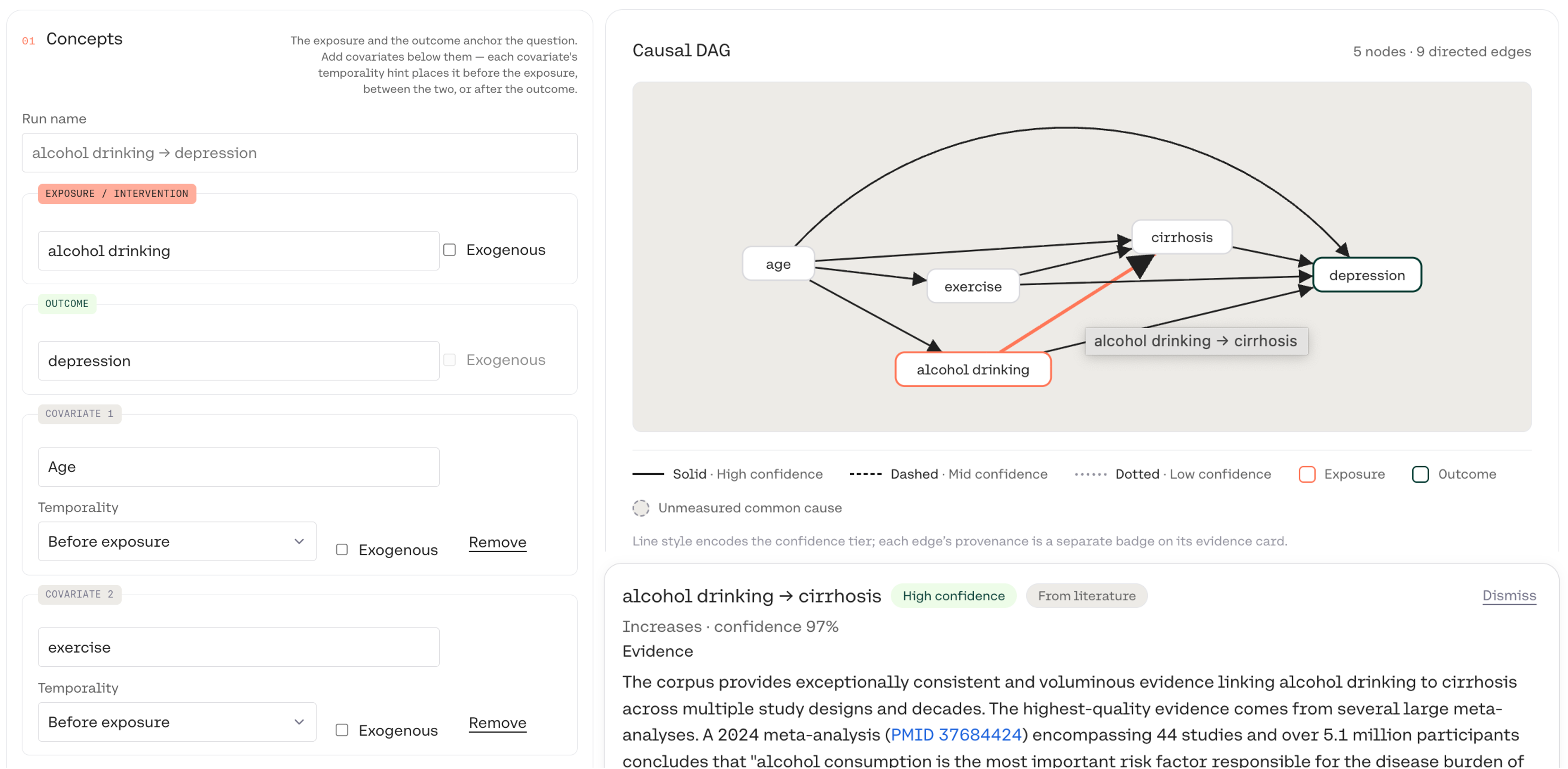}
\caption{\systemname{} interface. The left panel shows the study-submission form with temporal constraints. The right panel shows the completed causal DAG above a partially visible evidence card for the selected edge; the full card includes confidence, provenance, supporting citation, exact passage, and rationale.}
\label{fig:interface}
\end{figure*}

\paragraph{Study submission and review.}
Users specify an exposure or intervention, an outcome, covariates, and optional study context. They may add
temporal positions and required or forbidden edges as graph constraints. After validation, the run-detail
page tracks pipeline progress. At a configured review pause, \systemname{} stops before downstream consumption
so users can inspect concept mappings and retrieval coverage before continuing or revising.

\paragraph{Graph inspection and analysis.}
Completed runs open in an interactive DAG with user-readable node labels. Selecting an edge opens its
evidence card, which shows the proposed direction and confidence, supporting evidence channels and graph
decisions, cited PubMed records and exact passages when available, and a reviewable rationale. This makes
each edge's basis inspectable. Related views expose structural uncertainty through unexplained associations,
mediation decisions, and edge-level differences between alternatives. Users can compute backdoor adjustment
sets on demand and export the graph, evidence, and audit artifacts.

\paragraph{Evaluation console.}
A separate console runs evaluation batches through the same backend used for interactive studies. Users
select a reference cohort, one or more reference DAGs, and the number of replicate runs; the console
reports progress in real time and computes pairwise and graph-level metrics against the stored
references. Persisted configurations and results connect the interactive system directly to the
evaluation procedure reported in Section~\ref{sec:evaluation}.

\section{Experiments and Evaluation}
\label{sec:evaluation}

We evaluate \systemname{} as a system rather than as a single model call. Each benchmark item is converted
into an \systemname{} submission and run through the same staged backend used by the web interface. The
evaluation uses seven synthetic compact biomedical reference DAGs (S1--S7) and three larger literature
DAGs from published biomedical studies: A~\citep{boyle2015physical},
B~\citep{reiner2016machinery}, and C~\citep{evandt2017noise}. Notably, these references do not
necessarily reflect causal ground truth: each is one authored curation for a specific study, often
deliberately parsimonious, so structural precision, recall, and F1 measure reference reconstruction
rather than biological correctness. We report medians of three temperature-$0$ runs under the headline
\emph{all-tiers} policy, which scores every asserted edge independent of confidence tier. Results for the
high-only and high+mid confidence levels, along with configuration details and the scoring rationale,
appear in Appendix~\ref{app:reproducibility-details}. Each literature DAG is run in two input conditions: \emph{cold}, which
supplies only node labels, roles, and temporality anchors with no researcher-supplied study context,
and \emph{with study context}, which adds a serialized description of the source study's design; the
compact sDAGs are run cold.

Pairwise skeleton F1 scores the pairwise judgment table before graph assembly, ignoring direction and
asking only whether a concept pair is connected. Graph precision, recall, and directed F1 score the
final assembled DAG against the reference directed edge set.

Table~\ref{tab:benchmark-results} summarizes the main results. On the compact cohort, pairwise skeleton F1
is $0.944$ and graph-level directed F1 is $0.895$. Evidence grounding is strongest: across $66$ concept
pairs, every citation is a verbatim abstract span, and four pairs without literature are marked prior-only.

\begin{table}[t]
\small
\centering
\resizebox{\columnwidth}{!}{%
\begin{tabular}{lrrrr}
\toprule
Setting & Pairwise & \multicolumn{3}{c}{Graph} \\
\cmidrule(lr){2-2}\cmidrule(lr){3-5}
 & skel-F1 & P & R & dir-F1 \\
\midrule
sDAG mean & 0.944 & 0.835 & 0.980 & 0.895 \\
Literature mean, cold & 0.817 & 0.577 & 0.866 & 0.664 \\
Literature mean, w/ context & 0.843 & 0.625 & 0.905 & 0.714 \\
\bottomrule
\end{tabular}%
}
\caption{Aggregate benchmark results under the all-tiers inclusion policy (every asserted edge,
independent of confidence tier). Pairwise skeleton F1 scores pre-assembly connectedness between concept
pairs; graph metrics score the assembled DAG. The exposure--outcome estimand pair is excluded from both
metrics for the study-derived literature DAGs (whose design targets that edge) but retained for the
synthetic sDAGs (no study question); see the scoring rationale in Appendix~\ref{app:reproducibility-details}.
Values are medians of three temperature-$0$ runs. Higher precision, recall, and F1 are better.}
\label{tab:benchmark-results}
\end{table}

On the literature cohort (cold), mean graph directed F1 is $0.664$ at recall $0.866$ and precision
$0.577$: recall exceeds precision because \systemname{} asserts literature-supported relations that
parsimonious reference graphs may omit. Consistent with this interpretation, $86\%$ of literature-cohort
false-positive edges under the cold setting ($87\%$ with study context) carry a verified verbatim
citation extracted from a retrieved abstract, indicating that many apparent errors are reported
relations outside the reference DAG's chosen scope rather than unsupported model inventions. Study
context improves graph directed F1 from $0.664$ to $0.714$, precision from $0.577$ to $0.625$, and
recall from $0.866$ to $0.905$; the largest gain is on B, where context supplies \texttt{state} as an
exogenous anchor. A baseline that asks the same LLM to produce the graph directly better matches
parsimonious reference DAGs, but \systemname{} achieves higher recall on the literature cohort and uniquely provides
verifiable evidence trails for its surplus edges (Table~\ref{tab:llm-baselines}). Recovering more
reference-like parsimony without sacrificing per-edge grounding remains future work, as does a
controlled user study of the web interface.

\section{Related Work}
\label{sec:related-work}

\systemname{} links causal graph tooling, biomedical text mining, and retrieval-grounded language models.

\paragraph{Causal DAG tooling.}
Causal diagrams make identification assumptions explicit
\citep{greenland1999causal,pearl2009causality}. DAGitty supports drawing and adjustment-set analysis
\citep{textor2016dagitty}, while DoWhy supports graph-based identification, estimation, and refutation
once a graph has been specified \citep{sharma2020dowhy}. Causal discovery estimates structure from
observational or interventional data \citep{spirtes2000causation,zheng2018dags}. \systemname{} instead
addresses upstream authoring: producing an auditable candidate DAG from study variables and literature
before identification or estimation.

\paragraph{LLM-assisted causal graph construction.}
Recent work uses LLMs to answer causal-reasoning questions and supply background causal knowledge
\citep{kiciman2023causal}, provide structural priors to graph-discovery algorithms
\citep{darvariu2024llmpriors}, serve as conditional-independence oracles within constraint-based discovery
\citep{cohrs2024llmconstraint}, drive full causal-graph discovery through a breadth-first approach
\citep{jiralerspong2024efficient}, and combine domain-specific literature knowledge with data-driven
causal discovery \citep{barakati2025llmcausal}. \systemname{} integrates these roles in a bounded and auditable
way. Its LLM components make structured pairwise judgments from a fixed evidence snapshot and resolve only
explicitly specified graph ambiguities. Deterministic components enforce temporal ordering, user constraints,
and acyclicity. Mediation screening uses cited adjustment evidence when available, with a limited LLM fallback
when needed. Each edge includes confidence, provenance, a reviewable rationale, and citations when available.

\paragraph{Biomedical ontologies and literature infrastructure.}
Biomedical NLP draws on UMLS for concept normalization \citep{bodenreider2004umls}, PubTator for
automated concept annotation and search \citep{wei2019pubtator}, scispaCy for biomedical text processing
\citep{neumann2019scispacy}, and SciBERT and BioBERT for scientific and biomedical text representation
\citep{beltagy2019scibert,lee2020biobert}. SemRep extracts UMLS-grounded semantic predications, which
SemMedDB stores at PubMed scale \citep{rindflesch2003semrep,kilicoglu2012semmeddb}. BioCause annotates
causal relations in full-text biomedical articles \citep{mihaila2013biocause}, while SciFact links
scientific claims to supporting or refuting evidence and rationales \citep{wadden2020scifact}. These
resources support terminology normalization, literature processing, relation extraction, or claim-level
verification, but do not themselves construct study-specific causal DAGs. \systemname{} combines UMLS
enrichment with direct PubMed retrieval, freezes the retrieved corpus, and records query executions,
verification outcomes, and a PRISMA-inspired search summary \citep{page2021prisma}. It links these records
to pairwise judgments and graph decisions while assembling a constraint-checked primary DAG, retaining
uncertain relationships for review, generating alternatives when ambiguity remains, and supporting
on-demand identification analysis.

\paragraph{Retrieval-grounded LLM workflows.}
Retrieval-augmented generation combines a parametric generator with retrieved non-parametric memory for
knowledge-intensive NLP, allowing retrieved knowledge to be inspected alongside model outputs
\citep{lewis2020rag}. For causal graph recovery, Zhang et al.'s LACR retrieves scientific documents for
variable pairs, uses LLMs to assess conditional associations, and orients the resulting graph skeleton
\citep{zhang2024ragcausalgraph}. \systemname{} differs in its emphasis on evidence-preserving authoring: it
freezes a run-specific PubMed corpus, emits schema-constrained pairwise judgments, copies cited passages
directly from retrieved abstract snippets, deterministically enforces user constraints and acyclicity, and
records graph-level operations for review.

\section{Conclusion}
\label{sec:conclusion}

\systemname{} reframes automated causal DAG construction as evidence-preserving curation. Its main
contribution is a workflow in which LLMs make bounded pairwise and graph-level judgments over frozen
biomedical evidence, while deterministic constraints, provenance records, and expert review surfaces
keep the resulting assumptions inspectable. This framing is especially important for biomedical causal
analysis, where a useful system must expose why an edge was proposed, not merely produce a plausible
graph.

The evaluation highlights a key tradeoff between DAG construction approaches. The direct LLM
baseline's stronger structural F1 suggests that holistic graph framing can help match parsimonious
reference DAGs, while \systemname{}'s higher recall on the literature cohort and verified citations show the
value of staged, evidence-grounded curation. The next step is to make this staged workflow more
structure-aware: future versions should give pairwise judgments broader graph context and add
provenance-preserving review or pruning passes over assembled graphs.

\section*{Limitations}
\label{sec:limitations}

\systemname{} has several limitations. First, its evidence coverage is constrained: it relies on PubMed
abstracts rather than full text, UMLS coverage is uneven for specialized or temporally qualified
variables, and PubMed relevance ranking determines which records survive retrieval caps.

Second, LLM judgments remain fallible and model-dependent. Model priors may overstate familiar
relations, miss niche findings, or conflate total, direct, and mediated effects. Structured outputs,
citation grounding, deterministic constraints, and expert review make these errors inspectable but do
not guarantee causal correctness.

Finally, exhaustive concept-pair search and per-pair synthesis make retrieval and LLM cost grow
quadratically with variable count. Higher-dimensional use will require candidate-pair pruning, caching,
batching, and scalable graph interfaces.

\section*{Ethics and Broader Impact Statement}
\label{sec:ethics}

\systemname{} is intended as a research-support tool for drafting and auditing causal assumptions. Generated
DAGs do not establish causal validity and should not replace domain-expert reasoning. The interface keeps
experts in the loop by exposing concept mappings, retrieved evidence, pairwise judgments, graph operations,
and adjustment artifacts for review.

A central ethical risk is misplaced trust in fluent model-generated rationales. \systemname{} addresses
this by separating literature-grounded, ontology-grounded, prior-only, constraint-derived, and
solver-selected claims; preserving citations and audit records; and distinguishing low-evidence pairs
from confident null relationships. These mechanisms improve accountability, but they do not eliminate
the possibility of biased literature, incomplete retrieval, or erroneous LLM synthesis. We therefore
present generated DAGs as inspectable hypotheses for expert review, not as final causal conclusions.

\bibliography{bib/references}

\clearpage
\appendix
\section{Evaluation and Reproducibility Details}
\label{app:reproducibility-details}
\suppressfloats[t]

\paragraph{Standard evaluation configuration.}
All benchmark runs use a fixed \emph{standard} configuration. The provider/model is Anthropic
\texttt{claude-sonnet-4-6} at temperature~$0$; per-pair synthesis is parallelized under a concurrency
cap, but judgments are assembled in deterministic pair order. The default \emph{cold} setting provides
no researcher-supplied study context. Concept resolution uses UMLS enrichment. Literature retrieval uses
lexical query expansion and a per-DAG PubMed cutoff set to the day before the source study's online
publication date, preventing retrieval of the reference paper; synthetic sDAGs are unrestricted. Causal
synthesis produces one holistic judgment per pair over retrieved literature, UMLS relations, and model
prior knowledge, with citations selected from extracted snippets. DAG assembly uses the lenient mediation
policy, resolves supported associations as possible direct edges or candidate latent common causes,
and records an identification shell. Metrics are exported under high-only, high+mid, and
all-tiers inclusion policies. We report \textbf{all-tiers} as the headline because it evaluates every
asserted edge independently of confidence-tier calibration; Table~\ref{tab:policy-comparison} reports
all policies.

\paragraph{Full benchmark table.}
Table~\ref{tab:full-benchmark-results} gives the per-DAG results behind the aggregate rows of
Table~\ref{tab:benchmark-results}.

\begin{table}[t]
\small
\centering
\resizebox{\columnwidth}{!}{%
\begin{tabular}{lrrrr}
\toprule
DAG & Pairwise & \multicolumn{3}{c}{Graph} \\
\cmidrule(lr){2-2}\cmidrule(lr){3-5}
 & skel-F1 & P & R & dir-F1 \\
\midrule
\multicolumn{5}{l}{\textit{sDAG cohort (median of 3 runs, temperature 0)}} \\
S1 & 1.000 & 1.000 & 1.000 & 1.000 \\
S2 & 1.000 & 0.778 & 1.000 & 0.875 \\
S3 & 0.933 & 0.875 & 1.000 & 0.933 \\
S4 & 1.000 & 0.778 & 1.000 & 0.875 \\
S5 & 0.875 & 0.750 & 1.000 & 0.857 \\
S6 & 0.800 & 0.667 & 1.000 & 0.800 \\
S7 & 1.000 & 1.000 & 0.857 & 0.923 \\
\textbf{sDAG mean} & \textbf{0.944} & \textbf{0.835} & \textbf{0.980} & \textbf{0.895} \\
\midrule
\multicolumn{5}{l}{\textit{Literature cohort, cold (median of 3 runs, temperature 0)}} \\
A & 0.785 & 0.431 & 0.963 & 0.591 \\
B & 0.905 & 0.812 & 0.722 & 0.765 \\
C & 0.762 & 0.488 & 0.913 & 0.636 \\
\textbf{literature mean (cold)} & \textbf{0.817} & \textbf{0.577} & \textbf{0.866} & \textbf{0.664} \\
\midrule
\multicolumn{5}{l}{\textit{Literature cohort, with study context (median of 3 runs, temperature 0)}} \\
A & 0.792 & 0.455 & 0.926 & 0.610 \\
B & 0.930 & 0.842 & 0.833 & 0.811 \\
C & 0.805 & 0.579 & 0.957 & 0.721 \\
\textbf{literature mean (context)} & \textbf{0.843} & \textbf{0.625} & \textbf{0.905} & \textbf{0.714} \\
\bottomrule
\end{tabular}%
}
\caption{Full per-DAG benchmark results under the all-tiers inclusion policy. Pairwise skeleton F1
scores pre-assembly connectedness between concept pairs; graph metrics score the assembled DAG. Values
are medians of three temperature-$0$ runs. Higher precision, recall, and F1 are better.}
\label{tab:full-benchmark-results}
\end{table}

\paragraph{Inclusion-policy comparison.}
Confidence tiers determine which edges enter each inclusion policy. High-only and high+mid exclude
lower-confidence asserted edges, whereas all-tiers evaluates every asserted relation independently of
confidence calibration. This distinction affects the literature cohort but not the small cohort, whose
graphs carry no low-tier edges.

\begin{table}[t]
\small
\centering
\resizebox{\columnwidth}{!}{%
\begin{tabular}{lccc}
\toprule
Cohort & \multicolumn{3}{c}{Graph dir-F1} \\
\cmidrule(lr){2-4}
 & high-only & high+mid & all-tiers \\
\midrule
sDAG & 0.871 & 0.895 & 0.895 \\
Literature, cold & 0.486 & 0.623 & 0.664 \\
Literature, with context & 0.579 & 0.686 & 0.714 \\
\bottomrule
\end{tabular}%
}
\caption{Graph directed F1 under all three inclusion policies (cohort mean of per-DAG medians).
all-tiers is the headline; it includes edges assigned to lower confidence tiers that high-only and
high+mid exclude.}
\label{tab:policy-comparison}
\end{table}

\begin{table*}[t]
\small
\centering
\setlength{\tabcolsep}{4pt}
\begin{tabular}{lccccc}
\toprule
System & sDAG & Lit.\ cold & Lit.\ context & Verbatim & Audit \\
 & (P/R/F1) & (P/R/F1) & (P/R/F1) & citations & replay \\
\midrule
Sonnet 4.6, direct & 0.82/1.00/0.89 & 0.80/0.78/0.78 & 0.77/0.83/0.80 & 0 of 211 & no \\
Sonnet 4.6, pairwise & 0.76/0.63/0.68 & 0.67/0.76/0.70 & 0.70/0.58/0.63 & 0 of 812 & no \\
\textbf{\systemname{}} & \textbf{0.81/0.98/0.88} & \textbf{0.58/0.87/0.66} & \textbf{0.63/0.91/0.71} & \textbf{by construction} & \textbf{yes} \\
\bottomrule
\end{tabular}
\caption{\systemname{} vs.\ direct LLM baselines: graph directed precision / recall / F1 by cohort
(all-tiers; \systemname{} medians of 3, baselines single-pass at temperature~$0$), plus the traceability
contrast. On the literature cohort \systemname{} has the highest recall but lower precision (it asserts more
literature-supported edges the parsimonious references omit), while the direct baseline is more precise.
``Verbatim citations'' counts baseline citations whose quoted span is an exact substring of the cited
abstract, pooled across cohorts; none of the baselines' claimed citations are verifiable, whereas every
\systemname{} citation is an extracted-and-verified verbatim span.}
\label{tab:llm-baselines}
\end{table*}

\paragraph{LLM-only baselines.}
We compare \systemname{} to the same model (\texttt{claude-sonnet-4-6}, temperature~$0$) asked to produce a
DAG directly, with no retrieval, in two forms: \emph{direct} (one prompt returns the whole graph) and
\emph{pairwise} (one prompt per unordered pair, then a deterministic assembly). Both are scored through
the identical metric path (same reference, estimand/latent exclusions, all-tiers policy) as \systemname{}.
On literature graphs, \systemname{} trades precision for recall relative to the direct LLM baseline
(Table~\ref{tab:llm-baselines}). In the cold setting, \systemname{} reaches recall $0.87$ versus $0.78$ for
the direct baseline; with study context, recall is $0.91$ versus $0.83$. Its lower precision should be
interpreted against the construction of the references: these are parsimonious published DAGs, not
exhaustive biomedical knowledge graphs. Many \systemname{} false positives are therefore
literature-supported relations omitted from the reference graph, rather than unsupported model
inventions. This makes structural F1 alone an incomplete evaluation signal. The citation results differ
more clearly. Neither LLM baseline produced a citation whose quoted text appears verbatim in the cited
abstract (0 of 211 for direct; 0 of 812 for pairwise). The direct baseline names real PMIDs, but they
mostly point to the wrong papers, and its quotes do not appear in those abstracts. \systemname{} instead
extracts and verifies citations against the retrieved abstracts and retains the evidence associated with
each edge. A direct LLM may draw a plausible graph without providing a reviewable evidence trail.

\paragraph{Evidence grounding per DAG.}
Table~\ref{tab:grounding} separates pairwise judgments that could be tied to retrieved literature from
those that rely on model prior. For the compact graphs, almost every pair has abstract-backed support
($62$ of $66$). Coverage is more uneven in the literature cohort: cold runs ground $116$ of $161$
pairs, largely because B covers an occupational-injury setting with sparse PubMed evidence, where only
$11$ of $28$ pairs are citation-backed. \systemname{} keeps the remaining judgments but labels them
prior-only, so limited retrieval support remains visible. Exact CUI-to-CUI UMLS relations did not
ground any pair in these runs.

\begin{table}[t]
\small
\centering
\resizebox{\columnwidth}{!}{%
\begin{tabular}{lrrrr}
\toprule
Cohort / DAG & pairs & lit.-backed & prior-only & \% \\
\midrule
S1 & 6 & 6 & 0 & 100 \\
S2 & 10 & 10 & 0 & 100 \\
S3 & 10 & 9 & 1 & 90 \\
S4 & 10 & 10 & 0 & 100 \\
S5 & 10 & 7 & 3 & 70 \\
S6 & 10 & 10 & 0 & 100 \\
S7 & 10 & 10 & 0 & 100 \\
\textbf{sDAG total} & \textbf{66} & \textbf{62} & \textbf{4} & \textbf{94} \\
\midrule
A (cold) & 78 & 65 & 13 & 83 \\
B (cold) & 28 & 11 & 17 & 39 \\
C (cold) & 55 & 40 & 15 & 73 \\
\textbf{literature total (cold)} & \textbf{161} & \textbf{116} & \textbf{45} & \textbf{72} \\
\bottomrule
\end{tabular}%
}
\caption{Per-DAG evidence grounding, median of 3 runs. B has 28 pairs because its non-specific latent
is excluded; it relies heavily on model prior in its sparse-literature domain.}
\label{tab:grounding}
\end{table}

\paragraph{Scoring-rule rationale.}
The exposure--outcome estimand pair is excluded, in both orientations, from relation and graph metrics
for study-derived literature DAGs. These source studies were designed to estimate that effect, so
scoring it would conflate answering the research question with reconstructing the surrounding structure;
the pair is therefore dropped from system and reference edge sets and metric denominators. Synthetic
sDAGs are not drawn from studies, so their designated exposure/outcome pair is retained. We also exclude
non-specific latent placeholders such as $U^{*}$ and their incident edges, since \systemname{} reconstructs
latent confounders positionally rather than by unnamed-node identity. Named but unmeasured variables are
retained.

\paragraph{Error-analysis edges.}
Precision, not recall, is the principal structural gap, but the surplus edges are predominantly
literature-backed relations the parsimonious reference omits rather than model inventions: $86\%$ of
literature-cohort false-positive edges under the cold setting ($87\%$ with study context) carry a verified
verbatim citation extracted from a retrieved abstract. Some are transitively redundant; for example, the
S4 edge $\text{age}/\text{sex}\rightarrow\text{insulin resistance}$, which the reference routes through
obesity. Others are reported associations outside the reference DAG's chosen scope. In the literature
cohort, these concentrate in a few sink nodes (immune-function and vitamin-D for A; noise-annoyance,
sleep-medication, and chronic-disease for C).
Because each asserted edge carries its own verified citation, these can be inspected and accepted or
rejected individually, a distinction the LLM baselines cannot support, since none of their citations are
verifiable (0 of 211 direct, 0 of 812 pairwise).

\end{document}